\pdfoutput=1

\documentclass[11pt]{article}

\usepackage[final]{acl}

\usepackage{times}
\usepackage{latexsym}

\usepackage[T1]{fontenc}

\usepackage[utf8]{inputenc}

\usepackage{microtype}

\usepackage{inconsolata}

\usepackage{graphicx}

\usepackage{booktabs}

\newcommand{\unrel}{\rlap{\textsuperscript{*}}}

\usepackage[super]{nth}

\usepackage{float}
\usepackage{stfloats}

\usepackage{csquotes}

\title{Annif at SemEval-2025 Task 5: Traditional XMTC augmented by LLMs} 

\author{Osma Suominen \and Juho Inkinen \and Mona Lehtinen \\
        National Library of Finland, University of Helsinki \\ 
        \texttt{firstname.lastname@helsinki.fi}}

\begin{document}
\maketitle
\begin{abstract}

This paper presents the Annif system in SemEval-2025 Task 5 (LLMs4Subjects), which focussed on subject indexing using large language models (LLMs). The task required creating subject predictions for bibliographic records from the bilingual TIBKAT database using the GND subject vocabulary. Our approach combines traditional natural language processing and machine learning techniques implemented in the Annif toolkit with innovative LLM-based methods for translation and synthetic data generation, and merging predictions from monolingual models. The system ranked first in the all-subjects category and second in the tib-core-subjects category in the quantitative evaluation, and fourth in qualitative evaluations. These findings demonstrate the potential of combining traditional XMTC algorithms with modern LLM techniques to improve the accuracy and efficiency of subject indexing in multilingual contexts.


\end{abstract}

\section{Introduction}

Subject indexing is an important aspect of improving the discoverability of bibliographic databases and digital collections. Systems for automating subject indexing have traditionally been based on natural language processing (NLP) and traditional machine learning (ML) methods. The rise of generative AI and large language models (LLMs) holds some promise to revolutionise many automation tasks, yet producing accurate subject predictions using LLMs remains elusive (e.g. \cite{Chow03072024}, \cite{martins2024artificial}). The LLMs4Subjects challenge \cite{dsouza-EtAl:2025:SemEval2025} invited teams to produce innovative solutions for LLM-based subject indexing using a data set \cite{dsouza2024llms4subjects} based on the bilingual bibliographic database TIBKAT of TIB, the Leibniz Information Centre for Science and Technology.

We have been developing the Annif\footnote{\url{https://annif.org}} multilingual open source automated subject indexing toolkit since 2017 \cite{suominen2022annif}. Our toolkit is mainly based on traditional NLP and ML methods. By participating in this task, we aim to provide a strong baseline using traditional methods augmented with some LLM-based techniques. The novel aspects of our work are 1) translating the subject vocabulary and metadata records using LLMs, 2) generating synthetic training data using LLMs, 3) including XTransformer in an ensemble of algorithms, and 4) generating suggestions using separate monolingual prediction pipelines and merging their results.

Our system ranked \nth{1} in the \textit{all-subjects} category and \nth{2} in the \textit{tib-core-subjects} category in the quantitative evaluation. It was ranked \nth{4} in qualitative evaluations. We discovered that our approach, based mainly on traditional NLP and ML, remains competitive against other systems that make heavier use of LLMs. We also found ways to efficiently translate bibliographic records and to produce additional synthetic training data using LLMs.

Our code, configuration files and customised data sets are available on GitHub\footnote{\url{https://github.com/NatLibFi/Annif-LLMs4Subjects/}}. The models we trained are available on Hugging Face Hub\footnote{\url{https://huggingface.co/NatLibFi/Annif-LLMs4Subjects-data}}.

\section{Background}

The task involved developing LLM-based systems that recommend the most relevant subjects from the GND subject vocabulary to tag a given TIBKAT record based on its title and abstract, which is a type of extreme multilabel text classification (XMTC) problem. The organisers provided two variants of the GND subject vocabulary: \textit{all-subjects} (all 200,035 GND subjects) and \textit{tib-core-subjects} (a subset of 78,741 subjects especially important for TIB). We used the SKOS versions of GND provided by the German National Library (DNB).

The organisers also provided bibliographic records from the TIBKAT database as two data sets, corresponding to the two GND variants, each of them divided into train, development, and test subsets. The number of records was 81937 / 13666 / 27986 for the \textit{all-subjects} data set and 41923 / 7001 / 6119 for the \textit{tib-core-subjects} data set. All these subsets were further split by document type (e.g. Article or Book) and language (German or English). We did not distinguish records by these two aspects\footnote{We found that the records for different types were similar in their structure and their titles and abstracts often contained a mixture of languages regardless of the indicated language.}. Known GND subjects were included only for the train and development records.

Although the task description suggested using LLMs, we did not use one for the core task of choosing subjects but instead relied on the traditional XMTC algorithms in Annif. However, we used LLMs to pre-process the data sets and to generate additional synthetic training data.

\section{System overview}

We based our system on the Annif automated subject indexing toolkit. It provides a selection of XMTC algorithms as configurable \textit{backends} which can be used for subject indexing by setting up \textit{projects} that define the vocabulary and specific configuration settings of a backend.

We chose three Annif backends for this task: 1) \textbf{Omikuji}\footnote{\url{https://github.com/tomtung/omikuji}}, an implementation of a family of efficient machine learning algorithms for multilabel classification based on the idea of partitioned label trees, including Parabel \cite{prabhu2018parabel} and Bonsai \cite{khandagale2020bonsai}. We used the Bonsai-style configuration. 2) \textbf{MLLM}\footnote{\url{https://github.com/NatLibFi/Annif/wiki/Backend\%3A-MLLM}} (Maui-like Lexical Matching), a lexical algorithm for matching words and expressions in document text to terms in a subject vocabulary. It is a reimplementation of the ideas behind Maui \cite{medelyan2009human}, an earlier tool for automated subject indexing that uses heuristic features and a small machine learning model to select the best performing heuristics. 3) \textbf{XTransformer}, an XMTC and ranking algorithm based on fine-tuned BERT-style Transformer models that is part of the PECOS framework \cite{yu2022pecos}. Its Annif integration is experimental and was refined in the process of this task. We used \textbf{FacebookAI/xlm-roberta-base} as the base model.

Combinations of XMTC algorithms, called \textit{ensembles}, often outperform individual algorithms. We combined the base backends that return lists of suggested subjects along with numeric scores into two kinds of ensembles: \textit{simple ensembles} that merge subjects suggestions from two or more backends by averaging their scores, and \textit{neural ensembles} that, in addition to averaging scores, also involve training a neural network model that adjusts the subject scores, for example suppressing subjects that are frequently wrongly suggested (false positives).

\subsection{Translation of data sets}

We used the \textbf{Llama-3.1-8B-Instruct} LLM \cite{grattafiori2024llama} for translating the titles and abstracts of all records. Each record was translated separately into a German-only and English-only record (step 1 in Figure \ref{fig:preprocessing}). More details on LLM processing are given in Appendix \ref{app:llm_details}.

\begin{figure}[h]
  \includegraphics[width=\columnwidth]{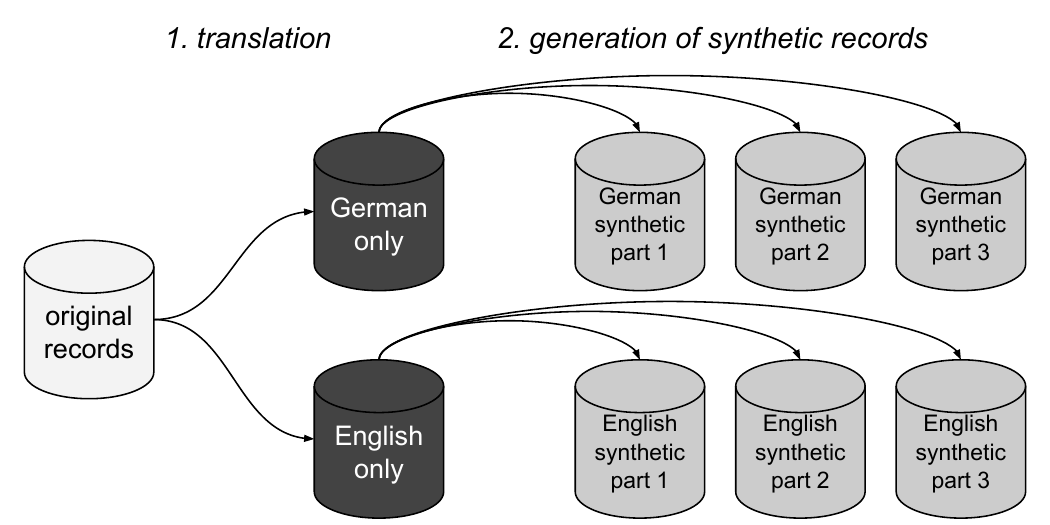}
  \caption{LLM pre-processing steps for the data sets.}
  \label{fig:preprocessing}
\end{figure}

The MLLM lexical backend requires that the vocabulary terms must be in the same language as the records. Therefore, we used the \textbf{GPT-4o-mini} LLM to translate all GND preferred terms in German into English and created bilingual variants of the GND SKOS files.

\subsection{Synthetic training data}

We found that the number of training records provided was quite small compared to the size of the subject vocabulary, so we used the \textbf{Llama-3.1-8B-Instruct} LLM to generate additional synthetic training records. We presented the LLM with each of the existing train records (its title and abstract) at a time along with its manually assigned subject labels (GND preferred terms in either German or English, matching the language of the document). We then asked the LLM to generate a similar record with the same set of subjects plus one additional, randomly chosen preferred term from the GND (step 2 in Figure \ref{fig:preprocessing}; see also example in Figure \ref{fig:llm-example} in Appendix \ref{app:llm_details}). This additional subject caused the LLM to generate a novel record, not just to rephrase the given example, and also helped to expand the subject coverage of the training data set to new GND subjects.

\section{Experimental setup}

We first set up, trained and evaluated the three kinds of base projects. We aimed to maximise their evaluation scores, measured against the development set, by exploring various approaches and settings. Once satisfied with the performance of the base projects, we combined them into ensembles that were finally used to produce our system output.

For evaluation during system development, we used two common XMTC metrics built in to the Annif toolkit: \textit{F1@5} (F1 score calculated using the top 5 suggestions from the system) and \textit{nDCG@10} (Normalised Discounted Cumulative Gain \cite{jarvelin2002}, a ranking metric calculated using the top 10 suggestions from the system).

\subsection{Base projects}

We set up parallel independent sets of Annif projects for the \textit{all-subjects} data set and the \textit{tib-core-subjects} data set. We also configured separate projects for English and German. To simplify the resulting combinatorial explosion of project configurations, we used the Data Version Control\footnote{\url{https://dvc.org/}} tool to manage the data sets, project configurations as well as the training and evaluation processes.

For each of the four combinations (2 GND variants \(\times\) 2 languages), we set up three Annif base projects: (Omikuji) Bonsai, MLLM and XTransformer (abbreviated as XTrans in tables). We trained each project on the LLM-translated monolingual records from the train set (German-only or English-only, matching the project language).

We also tested different hyperparameters. For each base project, we chose either Snowball stemming or Simplemma lemmatisation for text preprocessing. For the Bonsai projects we enabled bigram features using the \textit{ngram=2} setting and set a \textit{min\_df} value of 2 to 5 to filter features that occur rarely in the training data. For the XTransformer projects we manually searched for model-specific hyperparameters. The final hyperparameters can be seen in the project configuration files on GitHub\footnote{\url{https://github.com/NatLibFi/Annif-LLMs4Subjects/blob/main/projects.toml}}


\subsection{Adding synthetic data}

We repeated the synthetic data generation process three times per language and GND variant (total 3 \(\times\) 2 \(\times\) 2 times). The nDCG@10 scores of the Bonsai projects increased by \textasciitilde0.02 when adding the first part of synthetic data, but the \nth{2} and \nth{3} synthetic sets only increased the scores modestly (see Figure \ref{fig:synthetic-data} in Appendix \ref{app:effect_synthetic_data}), so we stopped at 1 part original and 3 parts synthetic data. We didn't use the synthetic data for training the MLLM and XTransformer projects. MLLM does not need much training data and the XTransformer results did not improve when adding synthetic data.

The final evaluation scores for the base projects are shown in Table \ref{tab:eval_base_projects} in Appendix \ref{app:eval_base_projects}. The Bonsai projects achieved the best scores in all cases, followed by XTransformer and MLLM.

\subsection{Ensemble projects}

\begin{figure}[h]
  \includegraphics[width=\columnwidth]{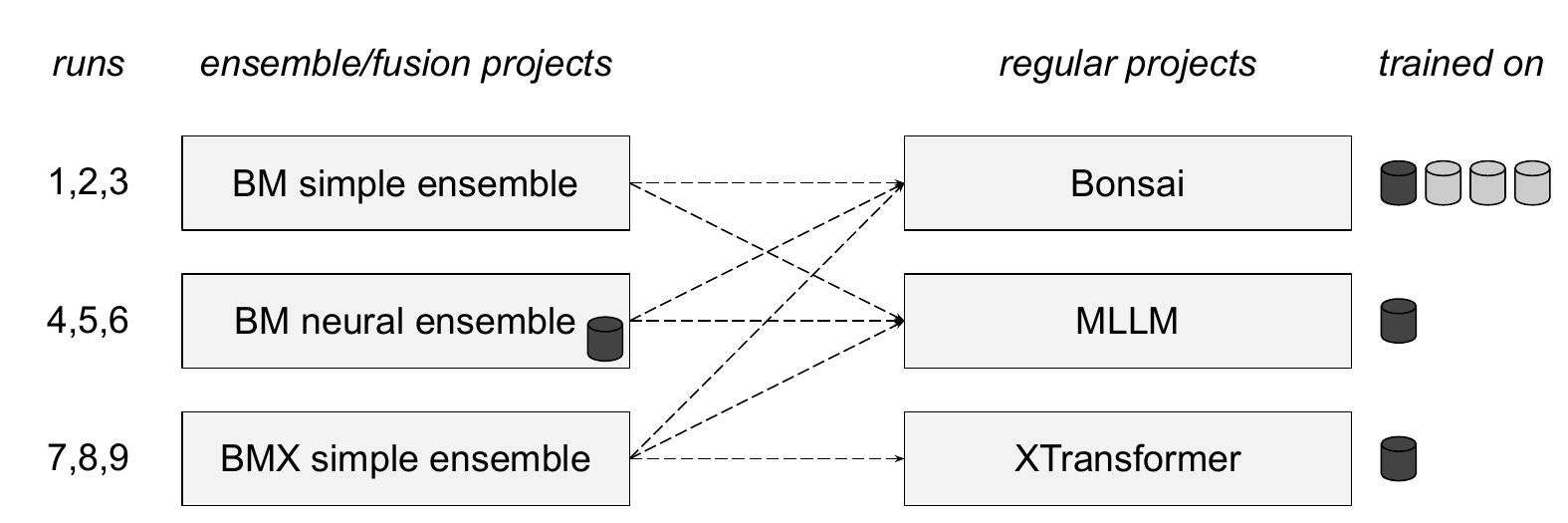}
  \caption{Overview of Annif projects and how they were combined into ensembles.}
  \label{fig:projects}
\end{figure}

We set up three kinds of ensemble projects that combined the base projects in different ways: two "BM" ensembles (simple and neural) combining Bonsai and MLLM, and a "BMX" simple ensemble that combines all three base projects (see Figure \ref{fig:projects}).

We tuned the ensembles using the automated hyperparameter optimisation facility built into Annif to select the weights used in averaging the results of base projects. We used the \texttt{annif hyperopt} command to try different combinations of weights (100 tries for the BM ensembles and 200 tries for the BMX ensembles), choosing weights that maximised the nDCG scores against the development set (see Table \ref{tab:ensemble_weights} in Appendix \ref{app:ensemble_weights}). In the BM ensembles, the Bonsai model contributed 80--87\% while MLLM had a minor role. In the BMX ensembles, the Bonsai model was still the most important at 47--62\%; XTransformer contributed 21--33\% while MLLM again had a minor role. These optimised weights match the relative order of the evaluation of the individual base projects, where Bonsai obtained the best results, followed by XTransformer and MLLM. Omikuji Bonsai and XTransformer are both similar models in the sense that they learn to recognise each subject individually based on the training data, but they are not good at suggesting concepts which occur with a low frequency. In the ensembles, MLLM complements these models by being able to suggest any subject in the GND vocabulary as long as the term used in the text matches the preferred or alternate label in the vocabulary.

We reused the optimised weights of the BM simple ensembles also for the corresponding BM neural ensembles and trained them for 10 epochs using records from the development set. The other NN ensemble hyperparameters were left at their default values.

The results of evaluating the ensembles against the development set are in Table \ref{tab:eval_ensemble_projects}. Since the BM neural ensembles were also trained on the development set records, their evaluation results were unrealistically good. As we had not set aside any other records for this purpose, we had to wait for the official evaluation results to assess their quality.

\begin{table*}[t]
  \centering
  \small
  \begin{tabular}{clcccccccc}
    \toprule
    \multicolumn{5}{c}{} & \multicolumn{2}{c}{development set} & \multicolumn{3}{c}{test set} \\
    \cmidrule(lr){6-7} \cmidrule(lr){8-10}
    Vocab & System & Run\# & Ensemble & Lang & F1@5 & nDCG@10 & F1@5 & Avg recall & Rank \\
    \midrule
    all & Annif (ours) &       1 & BM simple & de & 0.3174 & 0.5459 & 0.3108 & 0.5736 \\
    & &          2 &  & en & 0.3312 & 0.5677 & 0.3184 & 0.5890 \\
    & &          3 &  & de+en & - & - & 0.3376 & 0.6201 \\
    & &          4 & BM neural & de & 0.3337\unrel & 0.5726\unrel & 0.3029 & 0.5447 \\
    & &          5 &  & en & \textbf{0.3504}\unrel & \textbf{0.6008}\unrel & 0.3116 & 0.5599 \\
    & &          6 &  & de+en & - & - & 0.3318 & 0.6005 \\
    & &          7 & BMX simple & de & 0.3263 & 0.5614 & 0.3185 & 0.5859 \\
    & &          8 &  & en & 0.3411 & 0.5842 & 0.3276 & 0.6038 \\
    & &          9 &  & de+en & - & - & \textbf{0.3432} & \textbf{0.6295} & \nth{1} \\
    \cmidrule{2-10}
    & DUTIR831 &  3 &  & & & & 0.3346 & 0.6045 & \nth{2} \\
    & RUC Team &  1 &  & & & & 0.3015 & 0.5856 & \nth{3} \\
    & DNB-AI-Project &  1 &  & & & & 0.3231 & 0.5631 & \nth{4} \\
    & icip &  1 &  & & & & 0.2618 & 0.5302 & \nth{5} \\
    \midrule
    tib-core & Annif (ours) &  1 & BM simple & de & 0.2821 & 0.5557 & 0.2796 & 0.5285 \\
    & &          2 &  & en & 0.3009 & 0.5936 & 0.2984 & 0.5617 \\
    & &          3 &  & de+en & - & - & 0.3113 & 0.5824 \\
    & &          4 & BM neural & de & 0.3209\unrel & 0.6171\unrel & 0.2660 & 0.4864 \\
    & &          5 &  & en & \textbf{0.3467}\unrel & \textbf{0.6661}\unrel & 0.2886 & 0.5217 \\
    & &          6 &  & de+en & - & - & 0.3043 & 0.5559 \\
    & &          7 & BMX simple & de & 0.2891 & 0.5684 & 0.2864 & 0.5385 \\
    & &          8 &  & en & 0.3079 & 0.6051 & 0.3030 & 0.5719 \\
    & &          9 &  & de+en & - & - & 0.3136 & 0.5899 & \nth{2} \\
    \cmidrule{2-10}
    & RUC Team &  1 &  & & & & \textbf{0.3271} & \textbf{0.6568} & \nth{1} \\
    & LA\textsuperscript{2}I\textsuperscript{2}F &  2 &  & & & & 0.2717 & 0.5794 & \nth{3} \\
    & DUTIR831 &  2 &  & & & & 0.3153 & 0.5599 & \nth{4} \\
    & icip &  1 &  & & & & 0.2370 & 0.4976 & \nth{5} \\
    \bottomrule
  \end{tabular}
  \caption{Quantitative evaluation results for the ensemble projects measured against the development and test sets. Top 5 systems included for comparison. Note that Lang refers to the project language, not to the indicated language of the records.
  \small
  \textsuperscript{*}Unreliable score because the neural ensemble was trained on the development set it was evaluated on.}
  \label{tab:eval_ensemble_projects}
\end{table*}

\subsection{Test set predictions}

For both GND variants, teams were allowed to submit up to 10 separate \textit{runs} (up to 50 subject predictions per test set record). We LLM-translated the test set records to produce German-only and English-only versions of each record. For each of the three ensemble types, we produced three runs: 1) using the German ensemble and the German-only record, 2) using the English ensemble and English-only record, and 3) combining the two monolingual predictions into a single prediction by summing the scores of the predicted subjects and choosing the top 50 subjects by score. This gave us a total of 9 runs per GND variant that we submitted for evaluation.

\section{Results}

The final quantitative and qualitative evaluations were performed by the task organisers.

\subsection{Quantitative evaluation}

The quantitative evaluation involved comparing the subject predictions with subject annotations in TIBKAT records using precision, recall, and F1 scores (with various thresholds from 5 to 50). The overall ranking was determined by average recall, calculated by averaging the recall scores over all threshold values. Our \#9 runs, BMX simple ensemble with combined languages (de+en), ranked \nth{1} in the \textit{all-subjects} category with an average recall score of 0.6295 and \nth{2} in the \textit{tib-core-subjects} category with a score of 0.5899 (see Table \ref{tab:eval_ensemble_projects} for full results).

\subsection{Qualitative evaluation}

The qualitative evaluation was performed by subject librarians. Our \textit{tib-core-subjects} run \#9 was chosen for the qualitative evaluation. 6--10 record files from each of 14 different subject classifications were chosen, and the top 20 GND codes from the submissions were  evaluated by marking the predictions as correct (Y), technically correct but irrelevant (I), or incorrect (N or blank).

Based on these ratings, two different types of qualitative results were calculated. In case 1, both Y and I were considered correct, while in case 2, only Y was considered correct. Precision, recall and F1 scores across various thresholds (from 5 to 20) were calculated. For the recall calculation, the set of correct subjects was defined as the union of TIBKAT subject annotations and all the subject suggestions from the various systems that were considered correct by the evaluators. The systems were ranked based on their average recall scores across the specified thresholds. Our system ranked \nth{4} in both evaluations (see Table \ref{tab:qualitative_eval}).

\begin{table}[H]
    \centering
    \small
    \begin{tabular}{clcc}
        \toprule
        Case & System & Avg Recall & Rank \\
        \midrule
        1 & DNB-AI-Project & \textbf{0.5657} & \nth{1} \\
        & DUTIR831 & 0.5330 & \nth{2} \\
        & RUC Team & 0.5199 & \nth{3} \\
        & Annif (ours) & 0.5024 & \nth{4} \\
        & jim & 0.4928 & \nth{5} \\
        \midrule
        2 & DNB-AI-Project & \textbf{0.5094} & \nth{1} \\
        & DUTIR831 & 0.4851 & \nth{2} \\
        & RUC Team & 0.4645 & \nth{3} \\
        & Annif (ours) & 0.4484 & \nth{4} \\
        & jim & 0.4258 & \nth{5} \\
        \bottomrule
    \end{tabular}
    \caption{Qualitative evaluation results for the top 5 teams in evaluation cases 1 and 2.}
    \label{tab:qualitative_eval}
\end{table}

\subsection{Analysis}

We trained parallel English and German versions of each model, demonstrating successful LLM-based translation of multilingual input data. In all but one case, the English variant achieved higher evaluation scores than its German counterpart. The quality of LLM-produced translations can have an effect on the quality of the indexing of the downstream subjects. It may be that the translations were better in English or that the analytic structure of the English language makes it easier to process for traditional NLP pipelines than German, which is more synthetic and has many compound words. Further analysis of the effect of translation quality on the quality of subject indexing is left for future work.

By generating synthetic records, we were able to mitigate the lack of sufficient training data required by traditional ML algorithms. Thanks to this, the nDCG scores of our Bonsai models increased by \textasciitilde0.03 points (see Figure \ref{fig:synthetic-data} in Appendix \ref{app:effect_synthetic_data}).

The BMX ensembles consistently achieved higher evaluation scores than the corresponding BM ensembles, indicating that the addition of XTransformer had a positive effect. The neural BM ensembles achieved high evaluation scores against the development sets, as expected, but underperformed in the evaluations on the test set. In our experience, the neural ensemble is able to correct bias in settings where some of the training data are structurally different from the evaluation data. However, in this task, both the training and evaluation data was structurally similar so there was no need for such adjustment and the neural model simply made the predictions worse.

We tested a new "multilingual ensemble" method by generating predictions separately from German and English variants of the same records and then merging the subject predictions. The merged predictions achieved higher scores than the monolingual predictions. Had we not done this, our best runs would have been the BMX ensembles for English. In the quantitative evaluation, we would have ranked \nth{2} after DUTIR831 in \textit{all-subjects} and \nth{3} after LA\textsuperscript{2}I\textsuperscript{2}F in \textit{tib-core-subjects}. 

Our system ranked \nth{1} and \nth{2} in the quantitative evaluations, but in the qualitative evaluations, three other systems achieved higher scores. A possible explanation for this difference is that our system, based on traditional ML, was heavily guided by the training data records. We were thus able to produce subject predictions quite similar to the existing TIBKAT subject metadata. Some other systems, in contrast, produced predictions that were not as similar to existing metadata, but were considered qualitatively better by the evaluators. Assuming that the other systems relied more on LLMs for subject assignment than our system, they were not as constrained by the available training data and instead were able to leverage the knowledge of the LLMs. None of the systems achieved a F1@5 score above 0.35 in the quantitative evaluations, possibly indicating a relative lack of consistency in the TIBKAT subject metadata\footnote{In earlier experiments, we have been able to achieve F1@5 scores above 0.5 for some data sets that have been consistently indexed with good quality subject metadata.}.

Using different evaluation metrics for development and test sets introduced unnecessary complexity. In our opinion, the nDCG@50 metric would be a good choice for ranking systems that were tasked to produce 50 subject predictions per record.

\section{Conclusions}
In conclusion, our participation in the SemEval-2025 Task 5 (LLMs4Subjects) provided valuable insights into the capabilities and performance of Annif, particularly when augmented with large language models (LLMs) for data preparation. By leveraging traditional XMTC algorithms such as Omikuji Bonsai, MLLM, and XTransformer, and enhancing them with LLM-generated synthetic data and translations, we demonstrated competitive results across multiple categories. While this task focused only on the resulting quality of the subject indexing, we note that the computational requirements, energy consumption and processing latency of traditional ML approaches are modest in comparison to LLMs.

\section*{Acknowledgments}

The authors wish to thank the Finnish Computing Competence Infrastructure (FCCI) for supporting this project with computational and data storage resources. The authors thank the University of Helsinki Scientific Computing Services staff for all the support and assistance they gave for using the HPC environment. We thank ZBW, Leibniz Information Centre for Economics, for contributing the integration of XTransformer with Annif, and DNB for their valuable insight on its hyperparameters. Finally, we thank TIB for organising this task that provided valuable insights and opportunities for comparing methods and techniques.


\bibliography{custom}

\begin{thebibliography}{11}
\providecommand{\natexlab}[1]{#1}

\bibitem[{D'Souza et~al.(2024)D'Souza, Sadruddin, Israel, Begoin, and Slawig}]{dsouza2024llms4subjects}
Jennifer D'Souza, Sameer Sadruddin, Holger Israel, Mathias Begoin, and Diana Slawig. 2024.
\newblock \href {https://doi.org/10.5281/zenodo.15185475} {The {SemEval} 2025 {LLMs4Subjects} shared task dataset}.

\bibitem[{D'Souza et~al.(2025)D'Souza, Sadruddin, Israel, Begoin, and Slawig}]{dsouza-EtAl:2025:SemEval2025}
Jennifer D'Souza, Sameer Sadruddin, Holger Israel, Mathias Begoin, and Diana Slawig. 2025.
\newblock \href {https://aclanthology.org/2025.semeval2025-1.139} {{SemEval-2025} task 5: {LLMs4Subjects} - {LLM-based} automated subject tagging for a national technical library's open-access catalog}.
\newblock In \emph{Proceedings of the 19th International Workshop on Semantic Evaluation (SemEval-2025)}, pages 1082--1095, Vienna, Austria. Association for Computational Linguistics.

\bibitem[{Eric H. C.~Chow and Li(2024)}]{Chow03072024}
T.~J.~Kao Eric H. C.~Chow and Xiaoli Li. 2024.
\newblock \href {https://doi.org/10.1080/01639374.2024.2394516} {An experiment with the use of {ChatGPT} for {LCSH} subject assignment on electronic theses and dissertations}.
\newblock \emph{Cataloging \& Classification Quarterly}, 62(5):574--588.

\bibitem[{Grattafiori et~al.(2024)Grattafiori, Dubey, Jauhri, Pandey, Kadian, Al-Dahle, Letman, Mathur, Schelten, Vaughan et~al.}]{grattafiori2024llama}
Aaron Grattafiori, Abhimanyu Dubey, Abhinav Jauhri, Abhinav Pandey, Abhishek Kadian, Ahmad Al-Dahle, Aiesha Letman, Akhil Mathur, Alan Schelten, Alex Vaughan, et~al. 2024.
\newblock \href {https://arxiv.org/abs/2407.21783} {The {Llama} 3 herd of models}.
\newblock \emph{Preprint}, arXiv:2407.21783.

\bibitem[{J\"{a}rvelin and Kek\"{a}l\"{a}inen(2002)}]{jarvelin2002}
Kalervo J\"{a}rvelin and Jaana Kek\"{a}l\"{a}inen. 2002.
\newblock \href {https://doi.org/10.1145/582415.582418} {Cumulated gain-based evaluation of {IR} techniques}.
\newblock \emph{ACM Trans. Inf. Syst.}, 20(4):422–446.

\bibitem[{Khandagale et~al.(2020)Khandagale, Xiao, and Babbar}]{khandagale2020bonsai}
Sujay Khandagale, Han Xiao, and Rohit Babbar. 2020.
\newblock \href {https://doi.org/10.1007/s10994-020-05888-2} {Bonsai: Diverse and shallow trees for extreme multi-label classification}.
\newblock \emph{Machine Learning}, 109(11):2099--2119.

\bibitem[{Martins(2024)}]{martins2024artificial}
Sugabsen Martins. 2024.
\newblock \href {https://digitalcommons.unl.edu/libphilprac/8159/} {Artificial intelligence-assisted classification of library resources: The case of {Claude AI}}.
\newblock \emph{Library Philosophy and Practice}, 8159.

\bibitem[{Medelyan(2009)}]{medelyan2009human}
Olena Medelyan. 2009.
\newblock \href {https://hdl.handle.net/10289/3513} {\emph{Human-competitive automatic topic indexing}}.
\newblock Ph.D. thesis, The University of Waikato.

\bibitem[{Prabhu et~al.(2018)Prabhu, Kag, Harsola, Agrawal, and Varma}]{prabhu2018parabel}
Yashoteja Prabhu, Anil Kag, Shrutendra Harsola, Rahul Agrawal, and Manik Varma. 2018.
\newblock \href {https://doi.org/10.1145/3178876.3185998} {Parabel: Partitioned label trees for extreme classification with application to dynamic search advertising}.
\newblock WWW '18, page 993–1002, Republic and Canton of Geneva, CHE. International World Wide Web Conferences Steering Committee.

\bibitem[{Suominen et~al.(2022)Suominen, Inkinen, and Lehtinen}]{suominen2022annif}
Osma Suominen, Juho Inkinen, and Mona Lehtinen. 2022.
\newblock \href {https://doi.org/10.4403/jlis.it-12740} {Annif and {Finto AI}: Developing and implementing automated subject indexing}.
\newblock \emph{JLIS.it}, 13(1):265--282.

\bibitem[{Yu et~al.(2022)Yu, Zhong, Zhang, Chang, and Dhillon}]{yu2022pecos}
Hsiang-Fu Yu, Kai Zhong, Jiong Zhang, Wei-Cheng Chang, and Inderjit~S. Dhillon. 2022.
\newblock \href {https://arxiv.org/abs/2010.05878} {{PECOS}: Prediction for enormous and correlated output spaces}.
\newblock \emph{Preprint}, arXiv:2010.05878.

\end{thebibliography}

\appendix

\section{Effect of synthetic data}
\label{app:effect_synthetic_data}

\begin{figure}[H]
  \includegraphics[width=\columnwidth]{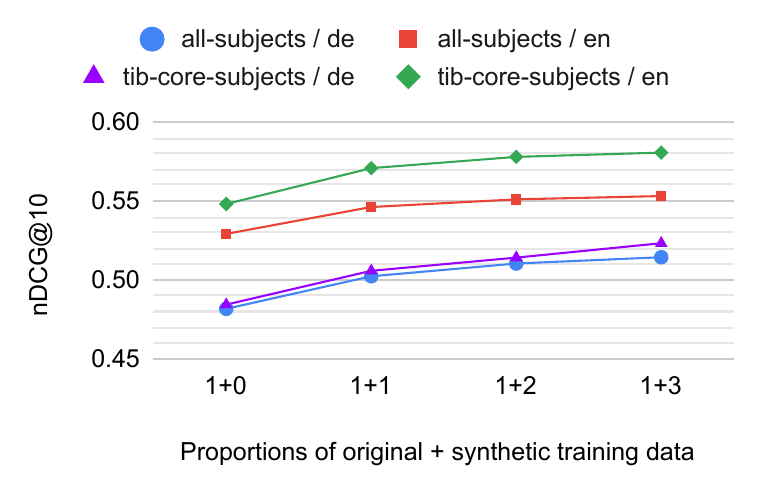}
  \caption{Effect of adding synthetic training data for Omikuji Bonsai models. The models were evaluated against the development set.}
  \label{fig:synthetic-data}
\end{figure}

\section{Results for base projects}
\label{app:eval_base_projects}

\begin{table}[H]
    \centering
    \small
    \begin{tabular}{ccccc}
        \toprule
        Vocab & Backend & Lang & F1@5 & nDCG@10 \\
        \midrule
        all & Bonsai &      de & \textbf{0.3003} & \textbf{0.5144} \\
         &  &               en & \textbf{0.3234} & \textbf{0.5532} \\
         & MLLM &           de & 0.2428 & 0.4245 \\
         &  &               en & 0.2281 & 0.4005 \\
         & XTrans &         de & 0.2928 & 0.5056 \\
         &  &               en & 0.3091 & 0.5326 \\
        \midrule
        tib-core & Bonsai & de & \textbf{0.2625} & \textbf{0.5233} \\
         &  &               en & \textbf{0.2934} & \textbf{0.5807} \\
         & MLLM &           de & 0.2087 & 0.4157 \\
         &  &               en & 0.2160 & 0.4291 \\
         & XTrans &         de & 0.2552 & 0.5052 \\
         &  &               en & 0.2761 & 0.5419 \\
        \bottomrule
    \end{tabular}
    \caption{Evaluation scores of the base projects measured against the development set. The best scores for each vocabulary and language have been set in \textbf{bold}.}
    \label{tab:eval_base_projects}
\end{table}

\section{Ensemble weights}
\label{app:ensemble_weights}

\begin{table}[H]
    \centering
    \small
    \begin{tabular}{cccccc}
        \toprule
        Type & Vocab & Lang & Bonsai & MLLM & XTrans \\
        \midrule
        BM & all & de & 0.8070 & 0.1930 & - \\
         &  & en & 0.8377 & 0.1623 & - \\
         & tib-core & de & 0.8432 & 0.1568 & - \\
         &  & en & 0.8729 & 0.1271 & - \\
        \midrule
        BMX & all & de & 0.4713 & 0.1964 & 0.3323 \\
         &  & en & 0.5387 & 0.1417 & 0.3196 \\
         & tib-core & de & 0.4891 & 0.1837 & 0.3272 \\
         &  & en & 0.6197 & 0.1671 & 0.2132 \\
        \bottomrule
    \end{tabular}
    \caption{Optimised weights of the base projects in the BM and BMX ensembles.}
    \label{tab:ensemble_weights}
\end{table}

\newpage

\section{Details about LLM usage}
\label{app:llm_details}

We used the vLLM\footnote{\url{https://docs.vllm.ai}} inference engine to translate and synthesise records using the Llama-3.1-8B-Instruct LLM. For the processing, we used a single NVIDIA A100 GPU with 80GB VRAM from the University of Helsinki HPC cluster Turso.

For translating GND into English, we used the GPT-4o-mini LLM on the Azure OpenAI Service cloud platform.

\subsection{Performance}

Using vLLM, we achieved a throughput of approximately 4 records/second for translation and 8 records/second for synthesising new records.

\subsection{Processing example}

\begin{figure*}[b]
  \includegraphics[width=\textwidth]{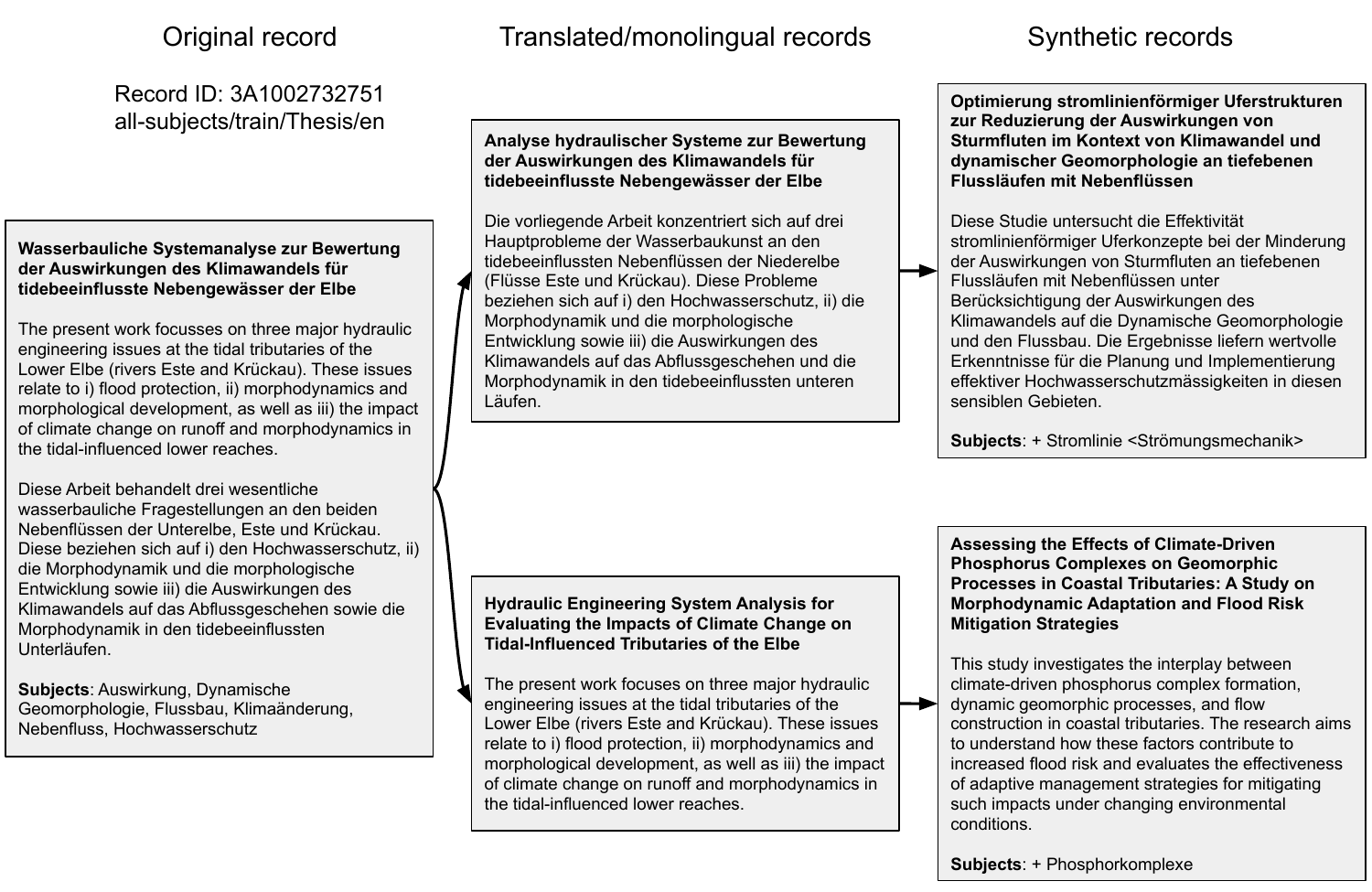}
  \caption{Example records translated and synthesised using the LLM. The example record included abstracts in both English and German, so the LLM only had to translate the German title to English. The LLM performed minor adjustment, for example changing "focusses" to "focuses" and modifying the German title. The synthetic records were generated using the translated records as one-shot examples but adding one random GND subject.}
  \label{fig:llm-example}
\end{figure*}

The process of LLM translation and record synthesis has been illustrated in Figure \ref{fig:llm-example}.

\subsection{Prompt templates}

In the following prompt templates, the system prompt is set in \textit{italics}. Tags in the template were replaced with relevant information from the records.

\subsubsection{Record translation}

This prompt was used to translate records:

\begingroup
\setlength{\parindent}{0pt}
\setlength{\parskip}{4pt} 
\raggedright
\small
\rule{\columnwidth}{0.5pt}
\textit{You are a professional translator specialized in translating bibliographic metadata.}

Your task is to ensure that the given document title and description are in <LANGUAGE> language, translating the text if necessary. If the text is already in <LANGUAGE>, do not change or summarize it, keep it all as it is.

Respond with only the text, nothing else.

Give this title and description in <LANGUAGE>:

<TITLE>

<DESCRIPTION> \\
\rule{\columnwidth}{0.5pt}
\endgroup

\subsubsection{Record synthesis}

This prompt was used to synthesise new records:

\begingroup
\setlength{\parindent}{0pt}
\setlength{\parskip}{4pt} 
\raggedright
\small
\rule{\columnwidth}{0.5pt}
\textit{You are a professional metadata manager.}

Your task is to create new bibliographic metadata: document titles and descriptions.

Here is an example document title and description in <LANGUAGE> with the following subject keywords: <OLD\_KEYWORDS>

<TITLE\_DESC>

Generate a new document title and description in <LANGUAGE>. Respond with only the title and description, nothing else. Create a new title and description that match the following subject keywords: <NEW\_KEYWORDS> \\
\rule{\columnwidth}{0.5pt}
\endgroup

\subsubsection{GND translation}

This prompt was used to translate the GND preferred terms into English in batches of 100 terms:

\begingroup
\setlength{\parindent}{0pt}
\setlength{\parskip}{4pt} 
\raggedright
\small
\rule{\columnwidth}{0.5pt}
\textit{You are a professional translator specialized in translating controlled vocabularies such as information retrieval thesauri and classifications.}

Your task is to translate terms from the The Gemeinsame Normdatei (GND, Integrated Authority File), a carefully curated thesaurus known for its precise and respectful terminology. These terms are used for academic and informational purposes and are presented in German. Please maintain the list structure and translate each term into English. Only return the list of translated terms, no explanations are needed.

This translation work is part of an educational and informational project aimed at enhancing accessibility and understanding of diverse concepts across languages. It is important to handle all terms, especially those pertaining to sensitive subjects such as health conditions, with accuracy and respect as intended by the thesaurus editors.

\vfill\eject

Example input:

1. Individualisierte Person \\
2. Familie \\
3. Schlagwort \\
4. Sicherung \\

Translated output for the above examples:

1. Differentiated person \\
2. Family \\
3. Subject heading \\
4. Safeguarding \\

Now translate the following thesaurus terms to English:

<LIST\_OF\_TERMS>
\rule{\columnwidth}{0.5pt}
\endgroup




\end{document}